\documentclass[11pt,  dvipsnames, table, x11name]{article}
\usepackage[toc,page]{appendix}
\usepackage[left=1in,top=1in,right=1in,bottom=1in, bindingoffset=0cm]{geometry}
\usepackage[font={small}, labelfont=bf]{caption}
\usepackage{subcaption}
\usepackage[utf8]{inputenc}
\usepackage{wrapfig}
\usepackage{xltabular}
\usepackage[LSBC3,T1]{fontenc}
\usepackage{fouriernc}

\usepackage{tikz}
\usepackage{tikzlings}
\usepackage{tikzpeople}
\usetikzlibrary{calc, shapes, backgrounds}
\usetikzlibrary{arrows, arrows.meta, calc, positioning}

\newdimen\nodeDist{}
\usepackage{pgfplots}

\pgfplotsset{compat=1.12}
\usepackage{blkarray}
\usepackage{comment}
\usepgfplotslibrary{fillbetween}
\usetikzlibrary{patterns}
\usetikzlibrary{quotes,angles}
\usetikzlibrary{positioning}
\usepackage{cancel}
\usepackage{booktabs}
\usepackage{array}
\newcolumntype{P}[1]{>{\centering\arraybackslash}p{#1}}
\usepackage{imakeidx}
\usepackage{fancyhdr}
\usepackage{graphicx}
\graphicspath{{./figures/}}
\usepackage{amsmath,commath,amssymb,amsthm,blkarray,bm}

\usepackage{mathtools}
\usepackage{titling}
\usepackage{mathrsfs}
\usepackage{physics}
\usepackage{enumitem}
\usepackage[toc,page]{appendix}
\usepackage{xcolor}
\definecolor{darkgreen}{RGB}{0,69,0} 
\definecolor{navy}{RGB}{0,60,113} 
\definecolor{backgroundcol}{RGB}{248, 249,250}
\definecolor{twc}{RGB}{66,92,133}
{\tiny } 
\definecolor{navy}{RGB}{0,60,113} 
\usepackage[pagebackref]{hyperref}
\hypersetup{
	colorlinks=true,
	linkcolor={blue!44!black},
	filecolor=magenta,      
	urlcolor=navy!75!black,
	citecolor={navy!75!black},
}
\newcommand{\R}{\mathbb{R}}

\newcommand\E[1]{\mathbb{E}\left(#1\right)}
\newcommand\Pro[1]{\mathbb{P}\left(#1\right)}

\definecolor{myorange}{RGB}{245,156,74}
\usepackage[authoryear]{natbib}
\usepackage{etoolbox}

\begin{document}
	\title{\textbf{Deep Learning for Causal Inference}\\ 
		\Large{A Comparison of Architectures for Heterogeneous Treatment Effect Estimation}}
	\author{Demetrios Papakostas\\ Andrew Herren \\P. Richard Hahn\\ Francisco Castillo}
	\date{}
	
	\maketitle
	\begin{abstract}
		Causal inference has gained much popularity in recent years, with interests ranging from academic, to industrial, to educational, and all in between.  Concurrently, the study and usage of neural networks has also grown profoundly (albeit at a far faster rate).  What we aim to do in this blog write-up is demonstrate a Neural Network causal inference architecture.  We develop a fully connected neural network implementation of the popular Bayesian Causal Forest algorithm, a state of the art tree based method for estimating heterogeneous treatment effects. We compare our implementation to existing neural network causal inference methodologies, showing improvements in performance in simulation settings.  We apply our method to a dataset examining the effect of stress on sleep.
	\end{abstract}
	
	\section{Introduction}
	A common problem in causal inference is to infer the effect of a binary treatment, 
	$Z$, on a scalar-valued outcome $Y$. When the effect of $Z$ on $Y$ is posited to be constant 
	for all subjects, or \textit{homogeneous}, the estimand of interest (average treatment effect) is a scalar-valued 
	parameter, which admits a number of common estimators. When the assumption of treatment effect homogeneity is 
	unwarranted, estimates of the average treatment effect (ATE) may be of questionable utility. The challenge of 
	estimating a heterogeneous conditional average treatment effect (CATE), is evident in the fact that the estimand is no longer 
	a scalar-valued parameter but a function of a (potentially high dimensional) covariate vector $X$. In recent years, 
	researchers have proposed to use machine learning methods for nonparametric CATE estimation (\cite{hahn2020bayesian}; \cite{krantsevich2021stochastic},
	\cite{hill2011bayesian}; \cite{wager2018estimation}; \cite{farrell2020deep}).  Additional methods that have been introduced include TARNET (\cite{shalit}), Dragonnet (\cite{shi2019adapting}) and that of \cite{chen2018heterogeneous}.  The main focus of this document will be comparing \cite{farrell2020deep} and the method we introduce, as they are the most similar in nature.  
	
	This paper focuses specifically on CATE estimators that rely on deep neural networks. 
	While neural networks are universal function approximators (\cite{cybenko1989approximation}), 
	nature does not typically provide treatment effects for use as ``training data," and estimation 
	proceeds by defining networks that can infer the CATE from available data.
	The architecture of a deep neural network, which refers to a specific composition of weights, data, and 
	activation functions, plays a crucial role in this process, along with regularization and training techniques. 
	This paper compares empirical CATE estimates of several architectures. The first two methods represent 
	outcomes as a sum of the CATE, $\beta(X)$, and the prognostic effect $\alpha(X)$, which occurs regardless 
	of treatment status. In the \cite{farrell2020deep} architecture, both $\alpha(X)$ and $\beta(X)$ emerge from a 
	shared set of hidden layers. Essentially, this architecture learns a common set of basis functions for 
	$\alpha(X)$ and $\beta(X)$ and then estimates separate coefficients for each those basis functions.
	We refer to this approach as ``the Farrell method" or simply ``Farrell" for the remainder of the paper.
	The second method an extension of Bayesian Causal Forests (BCF) (\cite{hahn2020bayesian}). This method, 
	which we hereafter refer to as ``BCF-nnet" or ``nnet BCF," uses completely separate neural networks for 
	$\alpha(X)$ and $\beta(X)$. Finally, we consider a ``naive" approach that partitions the data into 
	treatment and control groups, and learns a separate function on each subset of the data. These functions can be 
	used to estimate the CATE by subtracting predictions of the ``treatment function'' from those of the ``control function."
	
	Simulation studies show that nnet BCF outperforms both the Farrell and naive methods when 
	treatment effects are small relative to prognostic effects.
	\section{Problem Description}
	In order to make the problem precise, we begin by introducing notation and defining our estimators.
	We will use the following conventions in our notation:
	\begin{itemize}
		\item Bold upper case letters (i.e. $\mathbf{X}$) refer to random matrices
		\item Bold lower case letters (i.e. $\mathbf{x}$) refer to instantiations of random matrices
		\item Regular upper case letters ($X$, $Y$, $Z$) refer to random variables / vector
		\item Regular lower case letters ($x$, $y$, $z$) refer to instantiations of random variables
		\item Math calligraphy letters ($\mathcal{X}$, $\mathcal{Y}$, $\mathcal{Z}$) refer to the 
		support of a random variable
	\end{itemize}
	For example, if $X \sim f(X)$, we could write $\E{X} = \int_{\mathcal{X}} x f(x) dx$.
	
	Causal inference is concerned with the effect of a treatment, which we denote as $Z$, on an 
	outcome $Y$. In general, both the treatment and outcome can be continuous, categorical, or binary. 
	For the purposes of this paper, we restrict our attention to the case of a binary treatment 
	($\mathcal{Z} = \left\{0,1\right\}$) and a continuous outcome ($\mathcal{Y} = \R$). 
	
	Our overview of the causal inference assumptions largely follows that of \cite{hernan2020causal}. 
	We are interested in inferring the effect of a treatment, or intervention, on an outcome when 
	nothing is changed except the treatment being administered. In experimental settings, 
	this causal interpretation is often provided by the study design (randomized controlled trial, 
	randomized block design, etc...). In many real-world scenarios, designing and conducting an 
	experiment would be impossible, unethical, or highly impractical. In such cases, investigators 
	are limited to using observational data (i.e. data that were not collected from a designed 
	experiment).
	
	To formalize the idea described above, we let $Y^1$ and $Y^0$ denote two counterfactual random variables, 
	where $Y^i$ indexes the random outcomes for cases in which $Z$ has been set equal to $i$. The counterfactual 
	nature of these random variables is important to underscore before we define assumptions and estimators. 
	These two random variables are often referred to as \textit{potential outcomes} (see for example \cite{hernan2020causal}).
	The variables are random because even in the counterfactual scenario in which only treatment $i$ has 
	been administered, $Y$ may potentially be influenced by other factors.
	
	We can define the average treatment effect (ATE) as $\beta = \E{Y^1 - Y^0} = \E{Y^1} - \E{Y^0}$. 
	In many classical statistical problems, inferring the difference of two random variables is as 
	straightforward as assessing the difference in the empirical means of samples of both random variables. 
	However, the counterfactual nature of $Y^1$ and $Y^0$ is such that for any observation, only one of the two 
	variables can be observed (subjects cannot receive both the treatment and control). We define the 
	observable random variable $Y = ZY^1 + (1 - Z)Y^0$. In order to use a dataset of independent and identically 
	distributed (iid) samples of $Z$ and $Y$ to estimate the ATE, we must make several \textit{identifying} 
	assumptions.
	\begin{enumerate}
		\item \textit{Exchangeability}: $Y^1, Y^0 \perp Z$
		\item \textit{Positivity}: $0 < P(Z = 1) < 1$
		\item \textit{Stable Unit Treatment Value Assumption (SUTVA)}: $Y_i^1, Y_i^0 \perp Y_j^1, Y_j^0$ for all $j \neq i$
	\end{enumerate}
	With these assumptions, we can use observed data to estimate the average treatment effect. 
	One common estimator is the ``inverse propensity weighted" (IPW) estimator, whose expectation is shown below to 
	be the ATE under the three assumptions of exchangeability, positivity, and SUTVA.
	\begin{align*}
		\frac{YZ}{p(Z = 1)} - \frac{Y(1-Z)}{1 - p(Z = 1)} &= \frac{\left(Y^1Z + Y^0(1 - Z)\right)Z}{p(Z = 1)} - \frac{\left(Y^1Z + Y^0(1 - Z)\right)(1-Z)}{1 - p(Z = 1)}\\
		&= \frac{Y^1Z^2 + Y^0(1 - Z)Z}{p(Z = 1)} - \frac{Y^1Z(1-Z) + Y^0(1 - Z)^2}{1 - p(Z = 1)}\\
		&= \frac{Y^1Z}{p(Z = 1)} - \frac{Y^0(1 - Z)}{1 - p(Z = 1)}\\
		\E{\frac{YZ}{p(Z = 1)} - \frac{Y(1-Z)}{1 - p(Z = 1)}} &= \E{\frac{YZ}{p(Z = 1)}} - \E{\frac{Y(1-Z)}{1 - p(Z = 1)}}\\
		&= \E{\frac{Y^1Z}{p(Z = 1)}} - \E{\frac{Y^0(1-Z)}{1 - p(Z = 1)}}\\
		&= \E{Y^1}\E{\frac{Z}{p(Z = 1)}} - \E{Y^0}\E{\frac{1-Z}{1 - p(Z = 1)}}\\
		&= \E{Y^1}\frac{p(Z = 1)}{p(Z = 1)} - \E{Y^0}\frac{1-p(Z = 1)}{1 - p(Z = 1)}\\
		&= \E{Y^1} - \E{Y^0}
	\end{align*}
	In practice, $P(Z = 1)$ is often estimated from the data, but as long as $\E{\hat{p}(Z = 1)} = p(Z = 1)$, the 
	estimator will still be unbiased.
	
	IPW is just one of many estimators of the average treatment effect. We refer the interested reader to 
	\cite{hernan2020causal} for more detail. We now introduce the random variable $X$ to denote a 
	vector of \textit{covariates} of the outcome (often referred to as ``features" in machine learning). 
	These covariates might include demographic variables, health markers measured before treatment 
	administration, survey variables measuring attitudes or preferences, and so on. 
	
	Consider a simple motivating example, in which $X$ is age, $Z$ is a blood 
	pressure medication, and $Y$ is blood pressure. Older patients are more likely to have high blood 
	pressure, so we would expect that $X$ and $Y$ are not independent. Older patients, who visit 
	the doctor more frequently, are also potentially more likely to be prescribed blood pressure medicine. 
	In this case, we would not expect $Y^1, Y^0 \perp Z$. Older patients are more likely to receive blood pressure 
	medicine and also more likely to have high blood pressure so that observing $Z = 1$ changes the distribution of 
	$Y^1$ and $Y^0$. 
	
	We can work around this limitation with a modified assumption, \textit{conditional exchangeability}: 
	$Y^1, Y^0 \perp Z \mid X$. In words, this states that, after we control for the effect of $X$ on treatment 
	assignment, the data satisfy exchangeability. Similarly, we no longer have that $P(Z = 1)$ is the same 
	for all subjects, so we modify the positivity assumption to hold that $0 < P(Z = 1 \mid X) < 1$.
	Under this set of assumptions, we define a new IPW estimator as 
	$\frac{YZ}{p(Z = 1 \mid X)} - \frac{Y(1-Z)}{1 - p(Z = 1 \mid X)}$ and can show that its expected 
	value is the ATE.
	\subsection{Conditional Average Treatment Effect (CATE)}
	With the notation in place, we proceed to the focus of this paper: estimating heterogeneous 
	treatment effects using deep learning. In the prior section, we introduced the average treatment effect 
	as an expected difference in potential outcomes across the entire support set of covariates. 
	Average treatment effects have a long history in the causal inference literature because they are 
	(relatively) straightforward to estimate and provide useful, intuitive information about the average 
	benefits (or harms) of an intervention.  
	
	Sometimes, however, the ATE masks a considerable degree of heterogeneity in the causal effects of 
	an intervention. Consider the everyday example of caffeine tolerance. Some people find that any level 
	of caffeine consumption at any time of day carries too many unpleasant effects, while others drink 
	espresso after a large dinner. While it may be possible to measure an average treatment effect of a 
	given dose of caffeine, the estimate collapses a range of individual treatment effects and may thus 
	not provide much clinical or practical insight.
	
	We define the \textit{Conditional Average Treatment Effect} (CATE) as $\E{Y^1 - Y^0 \mid X = x}$. 
	Intuitively, this defines a treatment effect for the conditional distribution of $Y^1$ and $Y^0$ in which $X = x$. 
	Note that with this modification we define not a single parameter $\beta$ but a function $X \longrightarrow \beta(X)$. 
	If $X$ is binary or categorical, this can be done empirically by partitioning the data into subsets 
	$\left\{x: x = s\right\}$ and then estimating the ATE on the subsets. But in general, with continuous $X$ 
	or simply a large number of categorical $X$ variables, this approach becomes impossible and $\beta(X)$ 
	must be estimated by fitting a model.
	
	A tempting and convenient first step in CATE estimation would be use a linear model for $\beta(X)$. 
	More recently, advances in computer speed and a growing recognition of the complexity of many 
	causal processes has spurred interest in nonparametric estimators of $\beta(X)$. 
	To name a few examples, \cite{hahn2020bayesian} and \cite{hill2011bayesian} use Bayesian tree ensembles, 
	\cite{wager2018estimation} use random forests, and \cite{farrell2020deep} use deep learning. 
	The focus of this paper will be to compare the method introduced in \cite{farrell2020deep} to a novel architecture 
	inspired by \cite{hahn2020bayesian} and a naive partition-based architecture.
	\subsection{Estimating CATE using Deep Learning}
	We adapt the notation of \cite{farrell2020deep} slightly to fit the conventions used above. 
	As in prior sections, our goal here is to estimate a causal effect of a binary treatment $Z$, on 
	a continuous outcome $Y$. Since we are interested in the effect's heterogeneity, we must 
	construct a model that will estimate $\E{Y^1 - Y^0 | X = x}$ for any $x$. Before discussing the 
	specific architecture, we introduce some more clarifying terminology and notation. This construction of 
	treatment effect heterogeneity follows that of \cite{hahn2020bayesian}. Consider the following model
	\begin{align*}
		Y &= \alpha\left(X\right) + \beta\left(X\right) Z + \varepsilon\\
		\varepsilon &\sim \mathcal{N}\left(0, \sigma_{\varepsilon} \right)\\
		Z &\sim \textrm{Bernoulli}\left(\pi(X)\right)
	\end{align*}
	In this case, $\beta\left(X\right)$ corresponds to the treatment effect function, which given the 
	assumptions in the prior section, can be written as $\E{Y \mid X, Z = 1} - \E{Y \mid X, Z = 0}$.
	$\alpha(X)$ corresponds to $\E{Y \mid X, Z = 0}$ which we refer to as the \textit{prognostic function}, 
	$\varepsilon$ is random noise, and $\pi(X) = \Pro{Z = 1 | X}$ which we refer to as the 
	\textit{propensity function}.
	
	Right now, $X$ refers to a (potentially large) vector of covariates that may be useful in 
	estimating heterogeneous treatment effects. But using the above notation, we can partition 
	$X$ into several categories:
	\begin{enumerate}
		\item \textbf{Prognostic} features impact $Y$ through $\alpha(X)$
		\item \textbf{Effect-modifying} features impact the outcome $Y$ through $\beta(X)$
		\item \textbf{Propensity} features impact the outcome $Y$ through $\pi(X)$
	\end{enumerate}
	For example if $\pi(X) = \sin(X_1) + \abs{X_3}$, we would say that $X_1$ and $X_3$ are 
	propensity variables but $X_2$, for example, is not. These categories are of course not 
	mutually exclusive, but can be made so by considering their combinations. We avoid the 
	complete factorial expansion of these three categories and instead define several combinations 
	that are of particular interest in methodological problems.
	\begin{enumerate}
		\item \textbf{Pure prognostic} variables are variables which only appear in the function $\alpha(X)$
		\item \textbf{Pure modifiers} are variables which only appear in the function $\beta(X)$
		\item \textbf{Pure propensity} variables are variables which only appear in the function $\pi(X)$
		\item \textbf{Confounders} are variables which appear in both $\pi(X)$ and $\alpha(X)$
	\end{enumerate}
	
	Before we proceed, we also introduce the concept of \textbf{targeted selection}, when 
	$\pi(X) = f(\alpha(X))$. Intuitively, this corresponds to a practice of assigning treatment to those who 
	are most likely to need it (because, for example, $\alpha(X) = \E{Y | X, Z = 0}$ would be high otherwise). 
	This is an extreme version of confounding, in which the entire prognostic function is an input to the 
	propensity function and thus to the assignment of treatment. As is discussed in depth in \cite{hahn2020bayesian}, 
	this phenomenon is both highly plausible in real-world settings and also vexing to many approaches to CATE estimation.
	
	The architecture of the model is discussed in depth in later sections, so here we simply note the high-level 
	differences between the Farrell method and nnet-BCF. \cite{farrell2020deep} propose to fit a model 
	$\E{Y} = \alpha\left(X\right) + \beta\left(X\right) Z$ using a neural network with two hidden layers to which map to two 
	separate output nodes: $\alpha\left(X\right)$ and $\beta\left(X\right)$. 
	\cite{hahn2020bayesian} fit a similar model using Bayesian Additive Regression Trees (BART) (\cite{chipman2010bart}), 
	with one key distinction. $\alpha\left(X\right)$ and $\beta\left(X\right) Z$ are fit as completely separate models with no information 
	shared during training. This is different from the \cite{farrell2020deep} approach as their method shares weights between 
	the $\alpha(X)$ and $\beta(X)$ functions via the first two hidden layers. BCF nnet follows the approach of 
	\cite{hahn2020bayesian} by training two completely separate neural networks for $\alpha(X)$ and $\beta(X)$. 
	Finally, the ``naive" method estimates $\E{Y \mid X, Z = 1}$ with one network and $\E{Y \mid X, Z = 0}$ 
	with another network so that $\beta(X)$ can be estimated as a difference between these networks' predictions.
	
	\section{Methods}
	In this section, we discuss in more detail how the CATE is estimated in each of the three deep learning methods proposed, as well as a linear model comparison.
	\subsection{Joint Training Architecture (Farrell/Shared Network)}
\begin{figure}[!httb]
		\centering	
		\includegraphics[width=12cm]{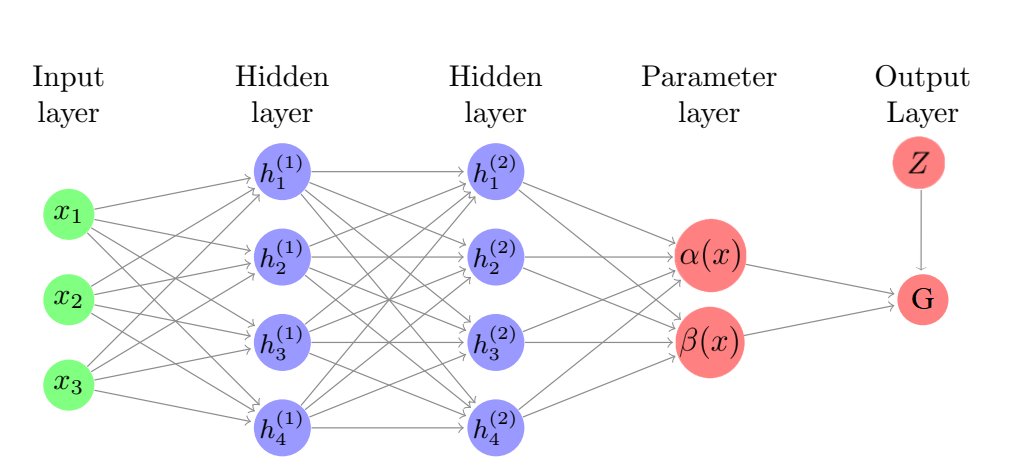}
		\caption{The Farrell method with a 3-dimensional vector of covariates $X$, 4 nodes in each hidden layer (in practice, these layers are usually much deeper). $G$ is an activation function that takes $\alpha(X)+\beta(X)Z$ as an argument.}
		\label{fig:farrell-pic}
	\end{figure}

	In \cite{farrell2020deep}, the authors posit that 
	\begin{equation}
		\mathbb{E}\left(Y\mid X=x, Z=z\right)=G\left(\alpha(x)+\beta(x)z\right)
		\label{farrell_eq}
	\end{equation}
	where $G(u), u\in \mathbb{R}$ is a known link function specified by the researcher, and $\alpha(\cdot)$ and $\beta(\cdot)$ are \emph{unknown} functions to be estimated. Since we are interested in effects of $Z$ on a real-valued $Y$, we use an identity link function so that $G()$ can be removed from the equations and we have $\mathbb{E}\left(Y\mid X=x, Z=z\right)=\alpha(x)+\beta(x)z$. The authors propose estimating $\alpha(\cdot)$ and $\beta(\cdot)$ with one deep fully connected neural network. We implement this architecture as a fully connected neural network with two hidden layers and a two-node parameter layer which outputs $\alpha(X)$ and $\beta(X)$. 
	The output of this architecture is then a linear combination of the two nodes in the parameters layer, $\alpha(x)+\beta(x)z$ (see Figure \ref{fig:farrell-pic}).
	
	Since $Y$ is real-valued, we use mean squared error (MSE) as a loss function in training each of the methods introduced in this section.
	
	\subsection{BCF nnet}
\begin{figure}[!httb]
		\centering
		\includegraphics[width=16cm]{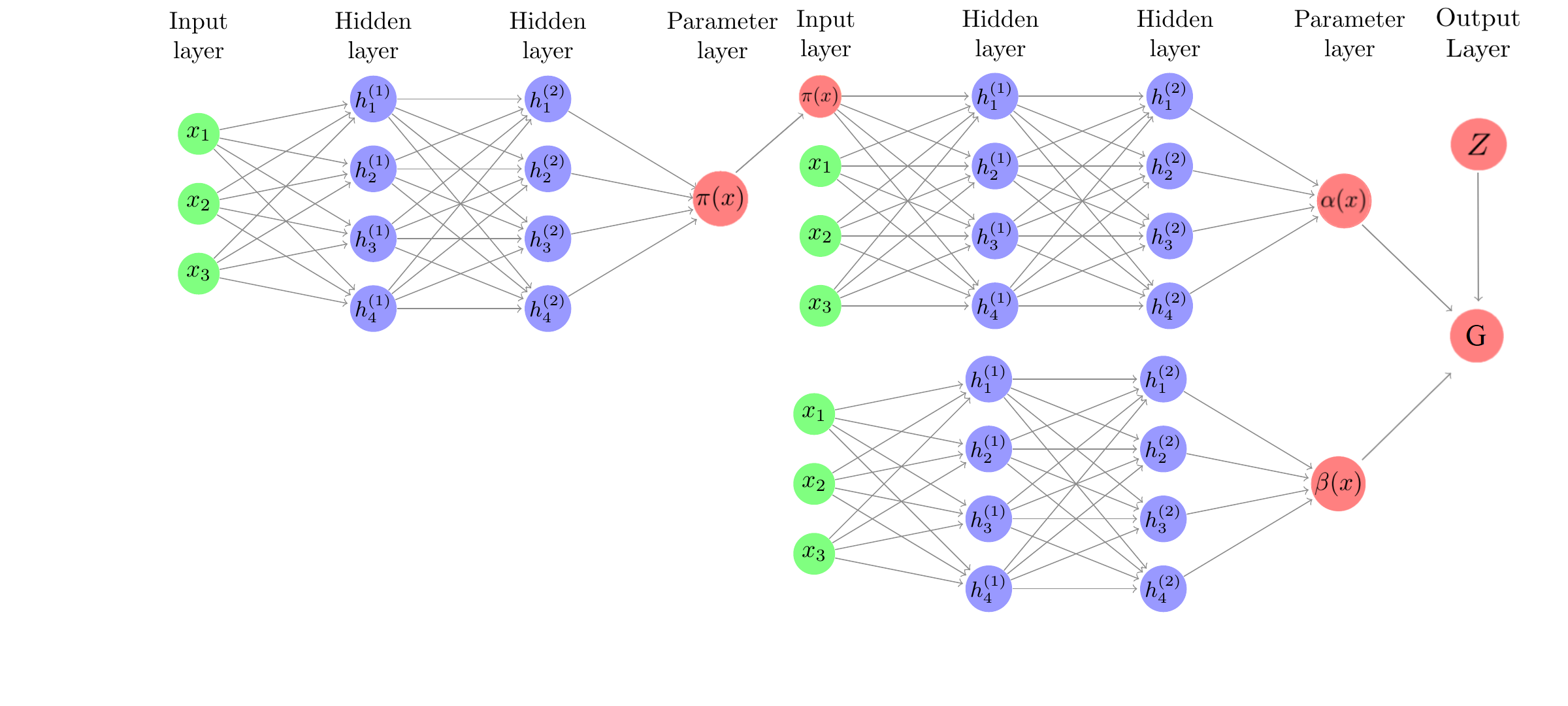}
		\caption{The BCF nnet architecture, where $G(\cdot)$ is an activation function that takes $\alpha(X)+\beta(X)Z$ as an argument.}
		\label{fig:nnbcf-pic}
	\end{figure}

	Based on the results and discussion in \cite{hahn2020bayesian}, we hypothesize that splitting $\alpha(\cdot)$ and $\beta(\cdot)$ 
	into separate networks with no shared weights may yield better CATE estimates on some data generating processes (DGPs). 
	The BCF nnet method specifies 
	\begin{equation}
		\mathbb{E}\left(Y\mid X=x, Z=z\right)=\alpha\left(x, \hat{\pi}(x)\right)+\beta(x)z
		\label{fig:BCF_main}
	\end{equation}
	In \cite{hahn2020bayesian}, $\alpha$ and $\beta$ are given independent BART priors (\cite{chipman2010bart}). 
	$\hat{\pi}(x_i)$ is an estimate of the propensity function
	We implement the BCF nnet architecture as in \autoref{fig:nnbcf-pic}.
	While the shared-weights versus separate weights distinction between Farrell and BCF nnet has been made clear, a subtle difference between the architectures is that BCF nnet allows for an estimate of the propensity function to be incorporated as a feature in the $\alpha(X)$ network. Since targeted selection implies $\alpha(X)$ is a function of $\pi(X)$, this parameterization was observed to be helpful in \cite{hahn2020bayesian}.
	
	In \cite{farrell2020deep}, the authors develop confidence intervals for their architecture's estimates (relying on influence functions, a common tool for calculating standard errors in non-parametrics).  We incorporated these intervals into our architecture, but found that they were far too tight and exhibited poor coverage in the low $n$ settings we were studying.  We therefore do not report or comment further.  
	\subsection{Separate Network Regression Approach}
	The ``naive'' method in our comparison employs two completely separate regression models, \begin{equation}
Y_1(X) = \E{Y \mid Z=1, X} \text{   and   } Y_0(X) = \E{Y \mid Z=0, X}
\label{eq:naive}
\end{equation}
With these two regression functions, our estimate of $\beta(X)$ is simply $\beta(X) = Y_1(X) - Y_0(X)$.  Each $Y_i(X)$ is constructed as a 2-layer fully connected neural network, with the number of parameters chosen to be similar to the number chosen for the Farrell and the BCF architecture.  
	
	\subsection{Linear Model}
	We also compare our two neural network architectures to a simple linear model's estimate of $\beta$
	\begin{equation}
		Y=\beta Z+X\delta +\varepsilon
		\label{eq:linmod_eq}
	\end{equation}
	where $\beta$ is the coefficient of interest and represents the average treatment effect. The model is fit using ordinary least squares (OLS).  We allow for interaction effects between $X$ and $Z$.

	\section{Simulation Summary}
	\autoref{eq:dgp1} is the first DGP we run. We choose a complex function for $\alpha$ and strong targeted selection, and a simpler function for $\beta$ (which allows for heterogeneous effects) to illustrate the effect of targeted selection.  
	\begin{align}
		\begin{split}
			X_1, X_2, X_3&\sim N(0,1)\\
			X_4&\sim \text{binomial}(n=2, p=0.5)\\
			X_5 &\sim \text{Bern}(p=0.5)\\
			X &= \qty(X_1, X_2, X_3, X_4, X_5)\\
			\beta\qty(X) &= \begin{cases}
				0.20+0.5*X_1\cdot X_4& \text{small treatment to prognosis}\\
				5+0.5*X_1\cdot X_4& \text{large treatment to prognosis}\\
			\end{cases}\\
			\alpha\qty(X)&=0.5\cos\qty(2X_1)+0.95*\abs{X_3\cdot X_5}-0.2*X_2+1.5\\
			\pi(X) &= 0.70*\Phi\qty(\frac{\alpha(X)}{s(\alpha(X))}-3.5)+u/10+0.10\\
			u&\sim \text{uniform}(0,1)\\
			Y&= \alpha(X)+\beta(X)Z+\sigma\varepsilon\\
			\varepsilon &\sim N(0,1)\\
			\sigma  &= \text{sd}(\alpha(X))\cdot \kappa \\
			Z &\sim \text{Bern}(p=\pi(X))
		\end{split}
		\label{eq:dgp1}
	\end{align}

	We choose the total number of parameters in the Shared architecture to be about the same as the separate network ($\alpha$ + $\beta$ networks). In the Shared network, this means we have 100 hidden nodes in layer 1, and 26 in layer 2, meaning 3,280 total parameters. 
	
	In the BCF NNet architecture, we have 60 parameters in the $\alpha$ first layer, 32 hidden nodes in the second layer.  For the $\beta$ network, we have 30 and 20 hidden nodes respectively. This yields 3,226 total parameters.  For both methods, we use a learning rate of 0.001 with an Adam Optimizer, we use Sigmoid activation, binary cross entropy loss for the propensity, MSE for the other networks, ReLu activation (double check), 250 epochs, and a batch size  of 64.  The  dropout  rate is 0.25 in every layer.
	The propensity score for the BCF NNet architecture is estimated using a 2 layer fully connected neural network with 100 and 25 hidden nodes respectively, and the rest of the parameters the same as above.  
In the separate network approach, we build the architecture
	separately using a 2-layer fully constructed neural network  (for both $Y_1$ and $Y_0$, as described in \autoref{eq:naive}) infrastructure with 50 hidden nodes in layer 1 and 26 in layer 2 for both models.  This is a total of 3,306 parameters. The other hyperparameters are the same as the BCF NNet and Shared Network approach.

	\autoref{tab:sim_results_smalltreat} shows results using both the Shared Network approach \cite{farrell2020deep} and the BCF Nnet approach we present.  This table indicates some RIC which biases the Farrell approach. The method we propose also has additional flexibility in that the propensity estimate can be estimated with any method and passed in, it need not be a MLP approach.  Additionally, because we separate the networks, like in the original BCF paper \cite{hahn2020bayesian}, we can add additional regularization on the $\beta$ network\footnote{In the world of neural networks, this could entail changing dropout rates, implementing early stopping, or weight-decay, amongst other approaches. In general, an advantage of Neural Networks, particularly when using a well developed and maintained service like pyTorch is the ease in customizing one's model for one's needs.}. 
	
\begin{figure}[!httb]
		\centering 
		\includegraphics[width=\textwidth]{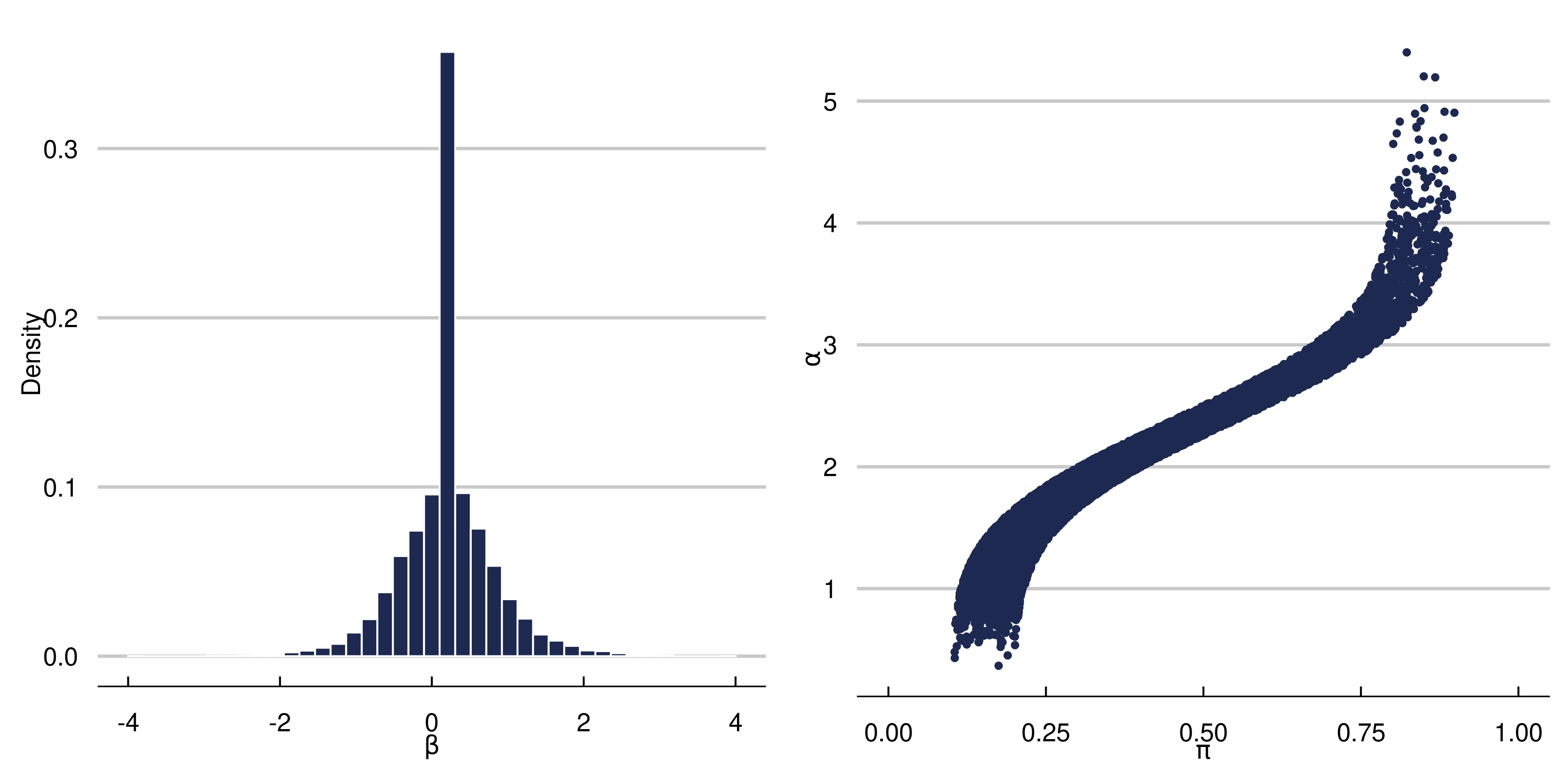}
		\caption{Left panel: Histogram of  $\beta$.  On the right is a plot of $\alpha$ vs $\pi$, indicative  of strong targeted selection.  For this particular realization of \autoref{eq:dgp1}, with $n=10,000$, the mean of $\beta(X)=0.20$, the mean of $\alpha(X)=1.95$, and the range of $\pi(X)=\qty(0.11, 0.90)$, with mean of 0.37.}
		\label{fig:dpg1}
	\end{figure}

	\autoref{tab:sim_results_largetreat} shows results with a large treatment to prognosis ratio.  In this setting, even with RIC presumably still being relevant due to the strong targeted selection in \autoref{eq:dgp1} (see right panel of \autoref{fig:dpg1}), the large treatment effect dominating allows for the extra parameters of the shared network approach to out-perform the separate network approach.  However, as the sample size  increases, the gap disappears, leading us to believe with sufficient sample size, this difference in methods would be minimal.

	\begin{table}[!httb]
	\centering
	\scalebox{.67}{
		\begin{tabular}{rrrrrrrrr}
			\toprule
			Method&n &  mean $\hat{\beta}$ &  True ATE&  True Mean $\alpha$ &  Mean Runtime &  Mean Correlation &  Mean rMSE &  mean (magnitude) Bias \\
			\midrule
			Shared Network& 250 &            0.47 &           0.20 &             1.95 &         45.38 &       0.74 &       0.50 &       0.26 \\
			BCF NNet&250 &        0.34 &           0.20 &             1.95 &         82.15 &       0.77 &       0.44 &       0.15 \\
			Separate Networks &250 &0.40 &           0.20 &             1.95 &         33.00 &       0.69 &       0.57 &       0.19 \\
			OLS Approach &250 & 2.04 &           0.20 &             1.95 &          0.00 &       0.16 &       2.00 &       1.83 \\ \midrule \midrule
			Shared Network & 500 &  0.38 &           0.20 &             1.95 &         47.85 &       0.80 &       0.42 &       0.18 \\
			BCF NNet&500 &          0.28 &           0.20 &             1.95 &         87.10 &       0.83 &       0.37 &       0.10 \\
			Separate Networks &500&    0.38 &           0.20 &             1.95 &         34.52 &       0.75 &       0.49 &       0.18 \\
			OLS Approach & 500 &         2.04 &           0.20 &             1.95 &          0.00 &       0.21 &       2.00 &       1.84 \\ \midrule \midrule
			Shared Network &1000 &            0.31 &           0.20 &             1.94 &         52.37 &       0.84 &       0.36 &       0.11 \\
			BCF NNet&1000 &            0.26 &           0.20 &             1.94 &         96.41 &       0.89 &       0.30 &       0.06 \\
			Separate Networks &1000 &    0.35 &           0.20 &             1.94 &         37.56 &       0.83 &       0.40 &       0.15 \\
			OLS Approach &1000 &       2.01 &           0.20 &             1.94 &          0.00 &       0.17 &       1.97 &       1.81 \\
			\bottomrule
		\end{tabular}
	}
	\caption{Simulation results with small treatment to prognostic ratio.  We run the DGP from \autoref{eq:dgp1} with $\kappa=1.0$ across 100 independent trials (varying $\varepsilon$) for each size $n$.  We generate $X$ for each size $n$, but keep the $X$ matrix the same for each of the 100 independent trials for each sample size (i.e. the $X$ vary across the sample sizes, but are constant at each iteration per sample size).  We train on the $n$ size, but test on a size of 10,000, to ensure we are looking at population parameters.  See \autoref{fig:BCF_better} if a visual presentation of this table is preferable. }
\label{tab:sim_results_smalltreat}
\end{table}

\begin{table}[!httb]
\centering
\scalebox{.67}{
	\begin{tabular}{rrrrrrrrr}
		\toprule
		Method&n &  mean $\hat{\beta}$ &  True ATE&  True Mean $\alpha$ &  Mean Runtime &  Mean Correlation &  Mean rMSE &  mean (magnitude) Bias \\
		\midrule
		Shared Network&250 &               4.89 &           5.00 &             1.94 &         46.00 &       0.60 &       0.62 &       0.14 \\
		BCF NNet&250 &           4.50 &           5.00 &             1.94 &         83.35 &       0.65 &       0.75 &       0.50 \\
		Separate Networks&250 &            4.85 &           5.00 &             1.94 &         33.49 &       0.49 &       0.87 &       0.19 \\
		OLS Approach&250 &                3.73 &           5.00 &             1.94 &          0.00 &       0.04 &       3.03 &      1.27 \\ \midrule \midrule
		Shared Network&500 &           4.95 &           4.99 &             1.95 &         48.20 &       0.66 &       0.51 &       0.09 \\
		BCF NNet&500 &         4.66 &           4.99 &             1.95 &         88.01 &       0.74 &       0.54 &       0.34 \\
		Separate Networks&500 &       5.01 &           4.99 &             1.95 &         34.90 &       0.49 &       0.72 &       0.10 \\
		OLS Approach&500 &          3.76 &           4.99 &             1.95 &          0.00 &       0.05 &       3.01 &      1.24 \\  \midrule \midrule
		Shared Network&1000 &     4.96 &           5.00 &             1.95 &         52.77 &       0.76 &       0.44 &       0.07  \\
		BCF NNet &1000 &        4.82 &           5.00 &             1.95 &         97.46 &       0.87 &       0.36 &       0.18 \\
		Separate Networks &1000 &            5.01 &           5.00 &             1.95 &         38.09 &       0.72 &       0.49 &       0.08  \\
		OLS Approach &1000 &        3.75 &           5.00 &             1.95 &          0.00 &       0.05 &       3.01 &      1.25 \\

		\bottomrule
	\end{tabular}
}
\caption{Simulation results with large treatment to prognostic ratio.  We run the DGP from \autoref{eq:dgp1} with $\kappa=1.0$ across 100 independent trials (varying $\varepsilon$) for each size $n$.  We generate $X$ for each size $n$, but keep the $X$ matrix the same for each of the 100 independent trials for each sample size (i.e. the $X$ vary across the sample sizes, but are constant at each iteration per sample size).  We train on the $n$ size, but test on a size of 10,000, to ensure we are looking at population parameters.  The benefits of the BCF NNet approach are lost, we hypothesize this is because any RIC introduced is offset by the large treatment, and since the treatment function compromises a good portion of the observed outcome, the extra parameters in the shared network should outperform the BCF NNet approach.  }
\label{tab:sim_results_largetreat}
\end{table}

\begin{figure}[!httb]
	\centering 
	\includegraphics[width=\textwidth]{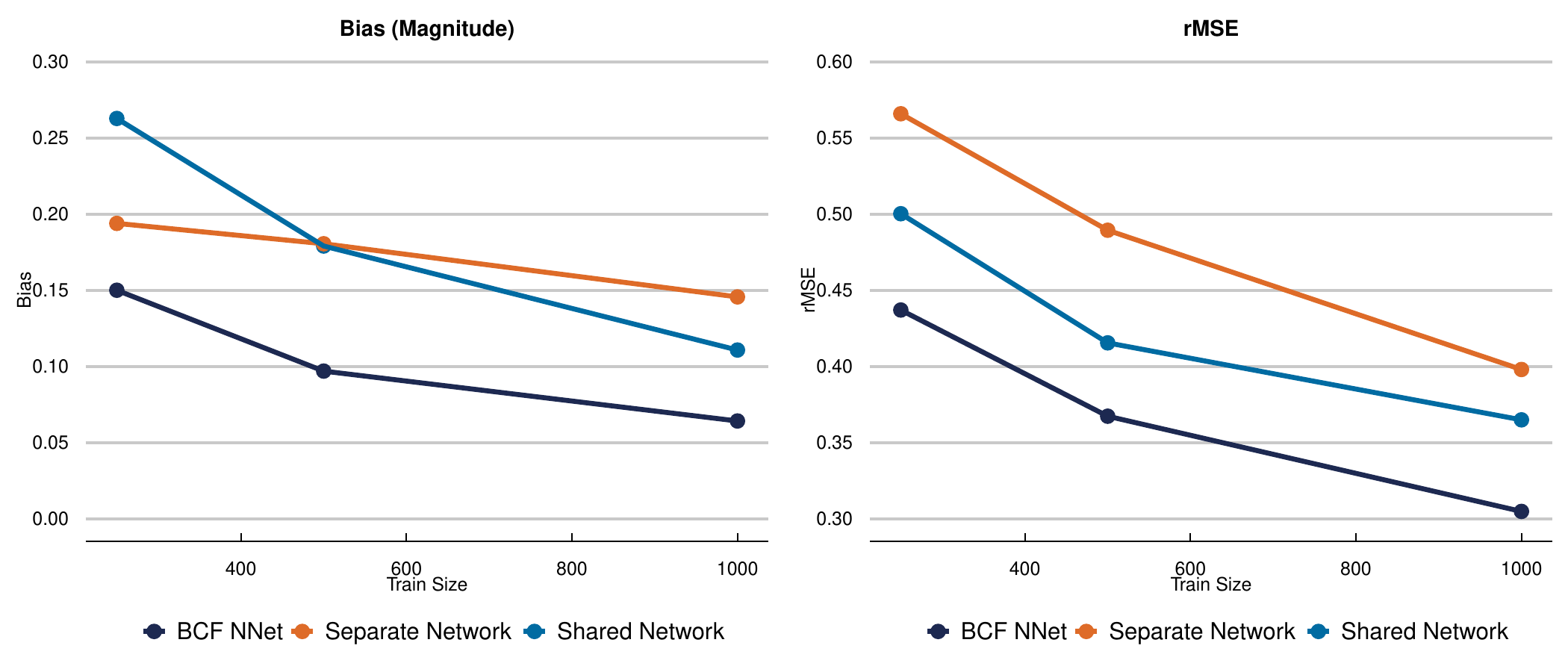}
	\caption{Left panel: Bias of dgp with different $n$. Right: RMSE.  This is in the ``small'' $\beta$ world.}
	\label{fig:BCF_better}
\end{figure}
\begin{figure}[!httb]
	\centering 
	\includegraphics[width=\textwidth]{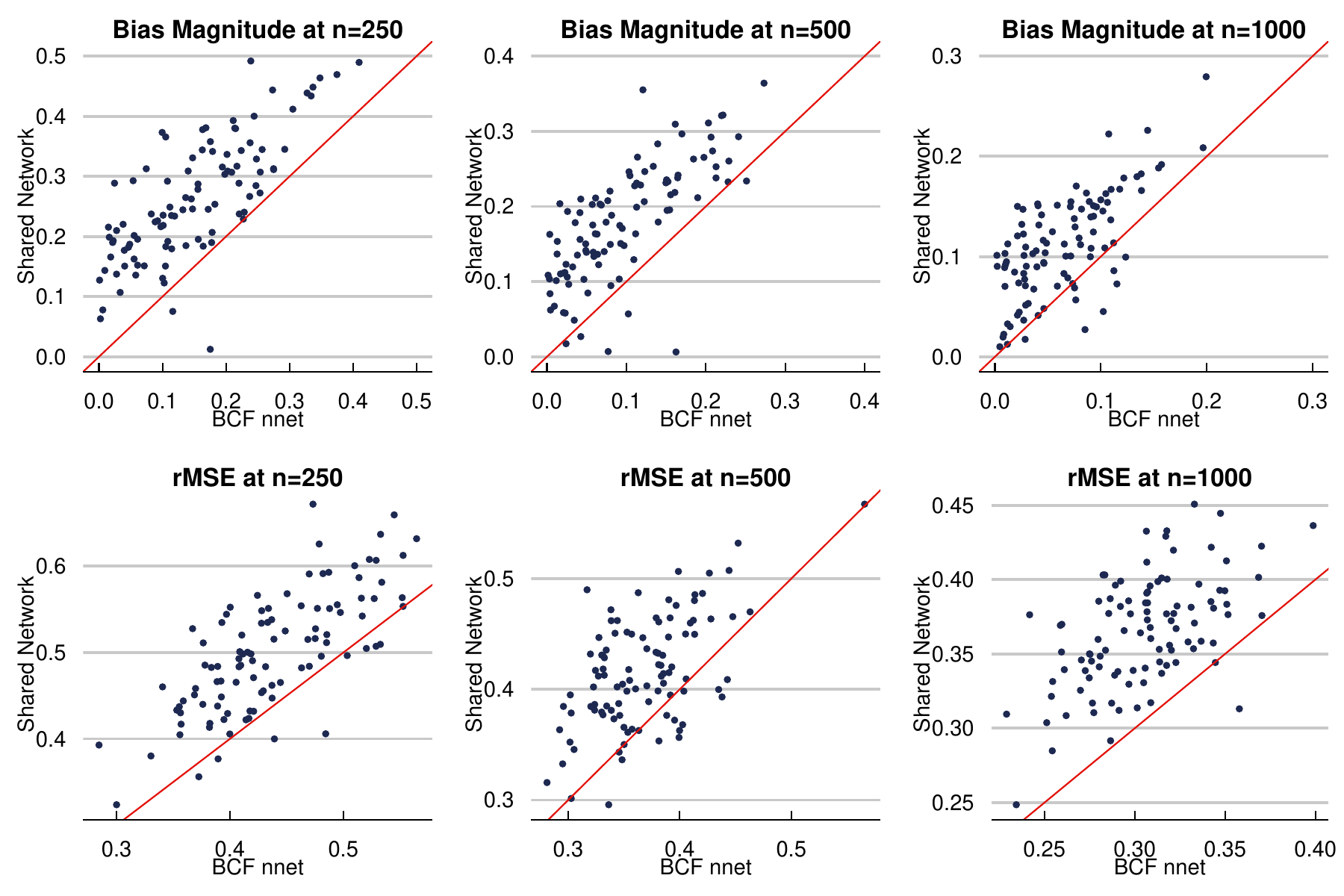}
	\caption{Comparing Individual biases and rmse's across the 100 Monte Carlo runs for the shared and BCF architectures.  }
	\label{fig:bcf_vs_shared}
\end{figure}
\clearpage

\section{Data Example}
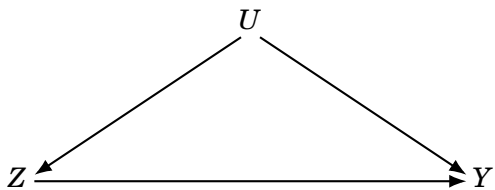
\begin{figure}[h]
	\centering
	\begin{tikzpicture}
		\node (X) at (0,0) {};
		\node at (-0.1,.05) {$Z$};
		\node at (3,2.15){$U$};
		\node (Z) at (3,2) {};
		\node (U) at (6,0) {}  ;
		\node at (6.1,.05) {$Y$};
		\draw[thick, -Latex](X)--(U)node [midway,below,sloped] {};
		\node at (4.5,1) [above, sloped] {};
		\node at (4.5,-1.5) [above,sloped] {};
		\draw [thick,-Latex] (Z) -- (X);
		\draw[thick,-Latex] (Z) -- (U);
	\end{tikzpicture}
	\caption[causal graph]{The classic graph.}
	\label{fig:data_explain}
\end{figure}

\cite{onyper2012class} is the data we look at as a demonstration of the methods.  The dataset we are interested in has 253 observations with 1 treatment variable, 1 outcome variable, and 11 variables defining our feature space $X$.  Specifically, the data is collected on participating undergraduate students at a liberal arts college in the northeastern United States. Our outcome is [{\tt Poor Sleep Quality}], which is a range of integers from 1-18, with 1 being the worst sleep quality and 18 being the best (the metric of interest is called the ``Pittsburgh Sleep Quality Index''),  We standardize the data using the  `scale' command in R.  The ``treatment'' we investigate is [{\tt Stress Level}], where students respond with ``high'' or ``normal''.  High stress level is considered the ``treatment'', and presumably should lead to lower sleep quality.  However, this problem is a very clear causal inference problem, as illustrated in \autoref{fig:data_explain}.  It is not unreasonable to expect there variables that could potentially lead to higher stress \emph{also} lead to worse sleep quality, and in particular those described below.  While this feature set likely does not fully meet the strong ignorability assumption, we  still muster on with the example. The more worrying assumption violation is the SUTVA violation.  

\begin{enumerate}
	\itemsep0em 
	\item {\tt Gender}: Male/Female
	\item {\tt Class Year}: Fresh, Soph, Jr, Sr.
	\item {\tt Early Class}: Whether or not the student signed up for a class starting before 9AM
	\item {\tt GPA}: College GPA, scale 0-4
	\item {\tt Classes Missed}: Number of classes missed in semester
	\item {\tt Anxiety Score}: Measure of degree of anxiety
	\item {\tt Depression Score}: Measure of degree of anxiety 
	\item{\tt Happiness Score}: Measure of degree of  happiness
	\item {\tt Number Drinks}: Number of alcoholic drinks per week
	\item{\tt All Nighter}: Binary indicator for whether student had all-nighter that semester
	\item{\tt Average Sleep}: Average hours of sleep for all days
\end{enumerate}

In our analysis, we change one hyper-parameter from our simulation study.  For the propensity estimation, we only run 100 epochs instead of 250 epochs.  In the simulation, we found that more epochs led to better predictive performance, but in the applied data problem, the propensity estimate clustered to estimates of 0 and 1 for the probability.  As a benchmark, we compared our fully connected 2-layer neural network estimate with a BART \cite{chipman2010bart} estimate, and found that with 100 epochs the estimates were similar.  Otherwise, all parameters stayed the same.  Additionally, for this analysis we do not do a train/test split.

\begin{table}[!httb]
	\centering
	\scalebox{.75}{
		\begin{tabular}{rrr}
			\toprule
			Method& CATE Estimate & Mean Prognostic \\
			\midrule
			
			Shared Network&1.13&NA\\
			Separate Networks&0.26&5.90  \\
			Separate Networks Bart Propensity &0.099 &6.07\\
			Naive NN Approach&-0.03& NA\\
			BCF (R-implementation)&0.07&6.20\\
			
			\bottomrule
		\end{tabular}
	}
	\caption{}
	\label{tab:data_result}
\end{table}

\begin{figure}
	\centering
	\begin{subfigure}{.5\textwidth}
		\includegraphics[width=\textwidth]{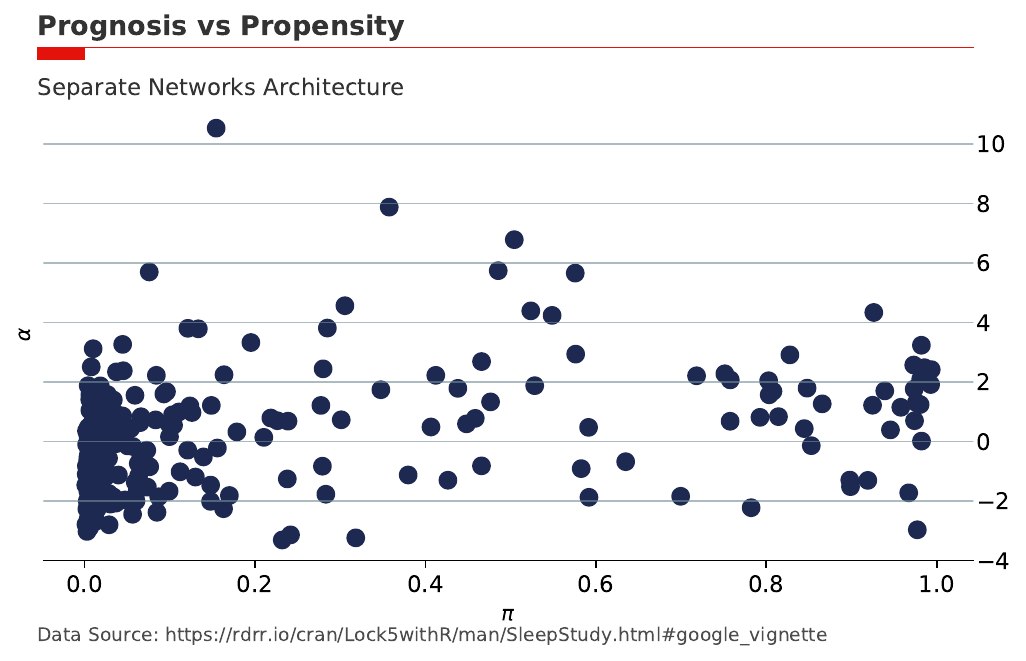}
	\end{subfigure}%
	\begin{subfigure}{.5\textwidth}
		\includegraphics[width=\textwidth]{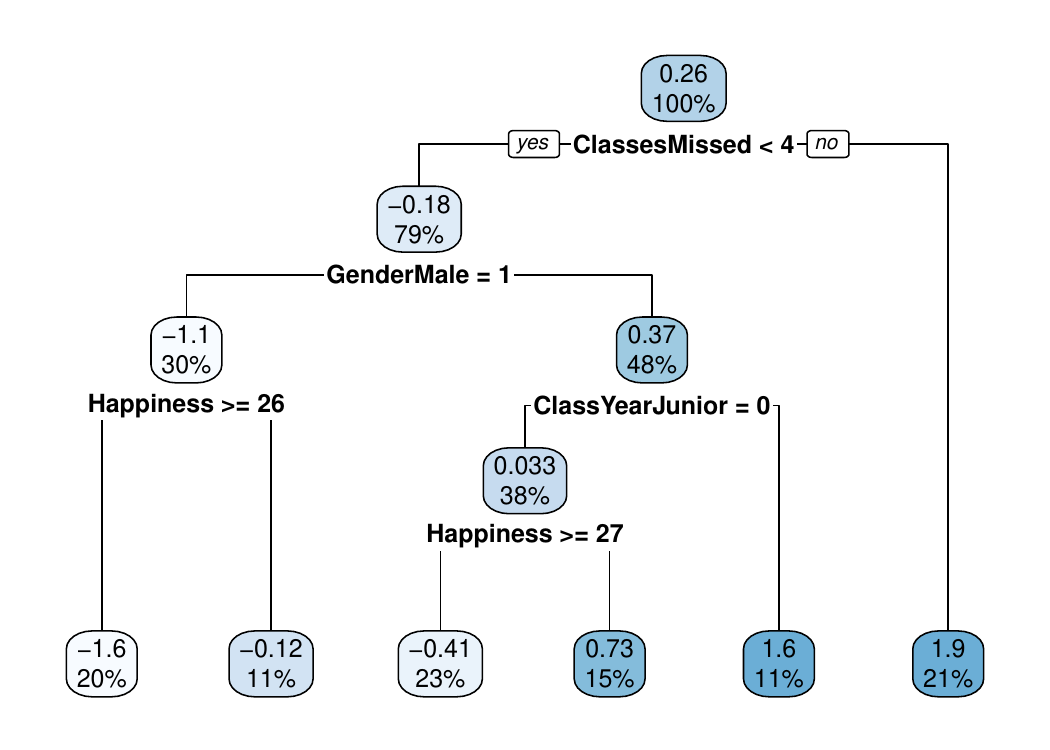}
	\end{subfigure}
	\caption{Left: A plot of $\alpha(X)$ vs $\pi(X)$, potentially indicative of targeted selection.  Of course, in this example the selection is not targeted, but one could convince themselves people may be more stressed out if they start thinking about sleep problems.  Right: A tree analysis, in the likes of \cite{Woody2020}.  We fit a tree to treatment effects and analyze the effects of moderating variables.}
\end{figure}

\section{Discussion}
What is preferable about this proposed methodology to the state of the art tree methods, such as \cite{wager2018estimation}, \cite{hahn2020bayesian}, or \cite{krantsevich2021stochastic}?  We do not aim to answer that question, but rather instead provide some evidence that if a researcher is intent on using some deep learning architecture for their causal needs,  then the methods developed in this document are the way to go.  For one, we show in a plausible simulation study the benefits of our methodology.  From a sheer performance point of view, our method provides an advantage over other competitive deep learning causal tools.  Additionally, the parameterization of \cite{hahn2020bayesian} provides multiple advantages.  Because we split the prognosis and treatment networks, we can regularize the networks differently, we could use different hyperparameters for each, we can include different controls for each.  This flexibility could likely be of importance to practitioners with expert knowledge.  Additionally, because the approach is built off neural networks, it lends itself to other applications, such as incorporating image data into the control set (or even as a treatment).  Building the network off pytorch also allows for easier scalability, adjustment, and online support.  
	\section*{Acknowledgements}
We thank ASU Research Computing facilities for providing computing resources.
\bibliography{deep-causal-paper.bib}

\begin{thebibliography}{13}
\providecommand{\natexlab}[1]{#1}
\providecommand{\url}[1]{\texttt{#1}}
\expandafter\ifx\csname urlstyle\endcsname\relax
  \providecommand{\doi}[1]{doi: #1}\else
  \providecommand{\doi}{doi: \begingroup \urlstyle{rm}\Url}\fi

\bibitem[Chen and Liu(2018)]{chen2018heterogeneous}
Ran Chen and Hanzhong Liu.
\newblock Heterogeneous treatment effect estimation through deep learning.
\newblock \emph{arXiv preprint arXiv:1810.11010}, 2018.

\bibitem[Chipman et~al.(2010)Chipman, George, McCulloch,
  et~al.]{chipman2010bart}
Hugh~A Chipman, Edward~I George, Robert~E McCulloch, et~al.
\newblock Bart: Bayesian additive regression trees.
\newblock \emph{The Annals of Applied Statistics}, 4\penalty0 (1):\penalty0
  266--298, 2010.

\bibitem[Cybenko(1989)]{cybenko1989approximation}
George Cybenko.
\newblock Approximation by superpositions of a sigmoidal function.
\newblock \emph{Mathematics of control, signals and systems}, 2\penalty0
  (4):\penalty0 303--314, 1989.

\bibitem[Farrell et~al.(2020)Farrell, Liang, and Misra]{farrell2020deep}
Max~H Farrell, Tengyuan Liang, and Sanjog Misra.
\newblock Deep learning for individual heterogeneity.
\newblock \emph{arXiv preprint arXiv:2010.14694}, 2020.

\bibitem[Hahn et~al.(2020)Hahn, Murray, and Carvalho]{hahn2020bayesian}
P~Richard Hahn, Jared~S Murray, and Carlos~M Carvalho.
\newblock Bayesian regression tree models for causal inference: Regularization,
  confounding, and heterogeneous effects (with discussion).
\newblock \emph{Bayesian Analysis}, 15\penalty0 (3):\penalty0 965--1056, 2020.

\bibitem[Hern{\'a}n and Robins(2020)]{hernan2020causal}
Miguel~A Hern{\'a}n and James~M Robins.
\newblock \emph{Causal Inference: What If}.
\newblock Boca Raton: Chapman \& Hall, CRC, 2020.

\bibitem[Hill(2011)]{hill2011bayesian}
Jennifer~L Hill.
\newblock Bayesian nonparametric modeling for causal inference.
\newblock \emph{Journal of Computational and Graphical Statistics}, 20\penalty0
  (1):\penalty0 217--240, 2011.

\bibitem[Krantsevich et~al.(2021)Krantsevich, He, and
  Hahn]{krantsevich2021stochastic}
Nikolay Krantsevich, Jingyu He, and P~Richard Hahn.
\newblock Stochastic tree ensembles for estimating heterogeneous effects.
\newblock \emph{Arxiv Preprint}, 2021.
\newblock URL \url{https://arxiv.org/abs/2209.06998}.

\bibitem[Onyper et~al.(2012)Onyper, Thacher, Gilbert, and
  Gradess]{onyper2012class}
Serge~V Onyper, Pamela~V Thacher, Jack~W Gilbert, and Samuel~G Gradess.
\newblock Class start times, sleep, and academic performance in college: a path
  analysis.
\newblock \emph{Chronobiology International}, 29\penalty0 (3):\penalty0
  318--335, 2012.

\bibitem[Shalit et~al.(2017)Shalit, Johansson, and Sontag]{shalit}
U.~Shalit, F.D. Johansson, and D.~Sontag.
\newblock Estimating individual treatment effect: Generalization bounds and
  algorithms.
\newblock \emph{International Conference on Machine Learning}, 2017.

\bibitem[Shi et~al.(2019)Shi, Blei, and Veitch]{shi2019adapting}
Claudia Shi, David Blei, and Victor Veitch.
\newblock Adapting neural networks for the estimation of treatment effects.
\newblock \emph{Advances in neural information processing systems}, 32, 2019.

\bibitem[Wager and Athey(2018)]{wager2018estimation}
Stefan Wager and Susan Athey.
\newblock Estimation and inference of heterogeneous treatment effects using
  random forests.
\newblock \emph{Journal of the American Statistical Association}, 113\penalty0
  (523):\penalty0 1228--1242, 2018.

\bibitem[Woody et~al.(2020)Woody, Hahn, and Murray]{Woody2020}
C.~Woody, S.~Carvalho, P.R. Hahn, and J.~Murray.
\newblock Estimating heterogeneous effects of continuous exposures using
  bayesian tree ensembles: revisiting the impact of abortion rates on crime.
\newblock \emph{Arxiv Preprint}, 2020.

\end{thebibliography}

\end{document}